\documentclass[pdflatex,final,3p,12pt]{elsarticle}
\usepackage{import}
\usepackage{dsfont}
\usepackage{url}
\usepackage{multirow} 
\usepackage{graphicx}
\usepackage{amsmath}
\usepackage{amsthm}
\usepackage{amssymb}
\usepackage{booktabs}
\usepackage{xcolor}

\begin{document}

\begin{frontmatter}

\title{MACMD: Multi-dilated Contextual Attention and Channel Mixer Decoding for Medical Image Segmentation }

\author[1]{Lalit Maurya\corref{cor1}}
\ead{lalit.maurya@port.ac.uk}
\author[1]{Honghai Liu}
\ead{honghai.liu@port.ac.uk}
\author[2]{Reyer Zwiggelaar}
\ead{rrz@aber.ac.uk}
\affiliation[1]{organization={School of Computing, University of Portsmouth},
            city={Portsmouth},
            postcode={PO1 3HE}, 
            country={UK}}
\affiliation[2]{organization={Department of Computer Science, Aberystwyth University},
            city={Aberystwyth},
            postcode={SY23 3DB}, 
            country={UK}}

\begin{abstract}
Medical image segmentation faces challenges due to variations in anatomical structures. While convolutional neural networks (CNNs) effectively capture local features, they struggle with modeling long-range dependencies. Transformers mitigate this issue with self-attention mechanisms but lack the ability to preserve local contextual information. State-of-the-art models primarily follow an encoder-decoder architecture, achieving notable success. However, two key limitations remain: (1) Shallow layers, which are closer to the input, capture fine-grained details but suffer from information loss as data propagates through deeper layers. (2) Inefficient integration of local details and global context between the encoder and decoder stages.
To address these challenges, we propose the MACMD-based decoder, which enhances attention mechanisms and facilitates channel mixing between encoder and decoder stages via skip connections. This design leverages hierarchical dilated convolutions, attention-driven modulation, and a  cross channel-mixing module to capture long-range dependencies while preserving local contextual details, essential for precise medical image segmentation. We evaluated our approach using multiple transformer encoders on both binary and multi-organ segmentation tasks. The results demonstrate that our method outperforms state-of-the-art approaches in terms of Dice score and computational efficiency, highlighting its effectiveness in achieving accurate and robust segmentation performance. The code available at \url{https://github.com/lalitmaurya47/MACMD}

\end{abstract}

\end{frontmatter}

\section{Introduction}
\label{sec:intro}
Medical image segmentation is essential in clinical workflows, enabling accurate disease diagnosis, effective treatment planning, and detailed quantitative analysis. In biomedical imaging, it is pivotal for identifying and delineating tumors, organs, and cellular structures despite challenges like variations in size, shape, and density. Deep learning methods \cite{lecun2015deep,Maurya2025} have revolutionized segmentation, outperforming traditional approaches. In MRI, segmentation enhances soft-tissue contrast for subtle abnormality detection, while in ultrasound, it delineates complex structures in real-time, improving diagnostic precision and optimizing therapeutic strategies efficiently.  
Traditional segmentation approaches have seen significant advancements with the introduction of CNNs by Long et al. \cite{long2015fully}, which have outperformed earlier methods in overcoming the challenges of medical image analysis. Furthermore, U-Net \cite{ronneberger2015u} has emerged as a foundational architecture, incorporating various advancements through its encoder-decoder framework and skip connections, which effectively integrate coarse and fine features to deliver exceptional segmentation precision \cite{qian2024multi}. Subsequent adaptations, incorporated attention mechanisms, dynamic modeling, and modified skip connections, enabling architectures like ResUNet \cite{diakogiannis2020resunet}, UNet++ \cite{zhou2018unet++}, UNet3+ \cite{huang2020unet}, and nnU-Net \cite{isensee2021nnu} to set new benchmarks. However, CNNs face inherent limitations in modeling long-range dependencies due to static receptive fields, which hinder their ability to capture broader spatial relationships critical for segmentation tasks. The introduction of attention mechanisms has improved feature refinement and pixel-level accuracy, but computational costs remain high, limiting scalability in resource-constrained environments.
Vision transformers (ViTs) \cite{dosovitskiy2020image} mark a paradigm shift, offering a global perspective on medical images by dividing them into non-overlapping patches embedded with positional information. ViT and its hierarchical variants such as Swin Transformer \cite{liu2021swin}, PVT \cite{ wang2022pvt}, and MaxViT \cite{tu2022maxvit}, among others, have been applied successfully to medical image segmentation tasks. These models excel in identifying correlations across spatial domains, addressing the limitations posed by CNNs in processing contextual information. However, ViTs encounter challenges in capturing local details, such as texture and structural nuances, which are equally vital for accurate segmentation in medical imaging.
Researchers have started integrating CNNs with ViTs to leverage their strengths in capturing both local and global features, resulting in innovative models such as TransUNet \cite{chen2021transunet}, MT-UNet \cite{wang2022mixed}, and HiFormer \cite{heidari2023hiformer}. These hybrid architectures excel in preserving fine details and enhancing segmentation performance. However, ViTs face challenges due to their self-attention mechanism, which has a computational complexity of $O(N^2)$. This makes them computationally intensive for high-resolution medical scans, posing a critical trade-off between accuracy and efficiency in clinical settings. Minimizing computational complexity while maintaining high accuracy is, therefore, highly desirable.

In medical image segmentation, both local texture details and global structural context are essential for accurate analysis. Skip connections play a vital role in this process by preserving high-resolution features and reintegrating scale information from early network layers into deeper layers, overcoming CNNs' naturally limited receptive fields \cite{azad2024medical}. In transformers, they enhance performance by directly incorporating local details into layers designed for global processing, balancing precision and contextual awareness \cite{ji2021multi}. However, most prior studies have primarily focused on refining the encoder or decoder, often neglecting innovations in skip connection design itself \cite{wang2022uctransnet, wang2024narrowing}. To address this gap, we propose the MACMD decoder architecture, which introduces enhanced skip connections incorporating four key modules. These modules enable multi-scale feature integration and attention-based refinement. They facilitate cross-channel mixing and spatial interplay between local and global features, thereby improving segmentation accuracy through context-rich feature reconstruction. The key contributions of this paper are summarized as follows.
\begin{itemize}
    
    \item  We present a novel decoding architecture that emphasizes innovations in skip connection design, enabling effective local-global context integration between encoder and decoder stages for medical image segmentation.
   
    \item We introduce four specialized modules in our decoder architecture, enhancing the attention mechanism and channel mixing to capture both fine local details and long-range contextual dependencies.
    
    \item We integrate MACMD with two standard encoders and demonstrate an improvement in segmentation performance across medical imaging modalities such as CT, ultrasound, and skin lesion analysis.
    
\end{itemize}

\section{Literature}
Skip connections have become a cornerstone in modern segmentation architectures, bridging the encoder and decoder stages to ensure effective information flow and mitigate the degradation problem that arises in deeper networks. By reusing encoder features in the decoder pathway, skip connections preserve essential spatial information, accelerate convergence, and promote gradient flow during training \cite{qian2024multi}. Inspired by residual connections in ResNets \cite{he2016deep} and the feature reuse strategy in DenseNets \cite{huang2017densely}, they facilitate the transfer of both semantic and geometric cues, which is crucial for refining object boundaries and localizing structures with higher precision. Advanced designs such as UNet++ \cite{zhou2018unet++} and UNet3+ \cite{huang2020unet} extend the concept further: UNet++ employs nested dense skip pathways to gradually bridge the semantic gap between encoder and decoder, while UNet3+ incorporates full-scale skip connections, enabling the integration of features across multiple resolutions. These designs enhance the network’s ability to capture fine-grained details alongside contextual information, resulting in improved segmentation accuracy.

Transformers have also been introduced to strengthen encoder–decoder communication. For example, UDTransNet \cite{wang2024narrowing} integrates a dual attention transformer that simultaneously models spatial and channel dependencies, ensuring that both structural layouts and feature importance are learned effectively. Similarly, MCTrans \cite{ji2021multi} emphasizes cross-scale contextual dependencies by combining self-attention and cross-attention modules. This allows the network to learn robust feature representations across varying scales and improve class correspondence, which is particularly important in handling heterogeneous medical imaging data.

Beyond architectural refinements, feature enhancement strategies play a vital role in boosting semantic richness and class discriminability. Techniques such as similarity matrices \cite{8361074} ensure intra-class coherence, sparse enhancement approaches \cite{9737529} reduce redundancy while emphasizing critical features, and context modeling \cite{9780146} enriches global dependencies for more holistic understanding. These methods collectively sharpen feature maps, making them more focused and reliable for segmentation tasks.

Complementary to enhancement, multi-scale feature aggregation has emerged as a powerful means to integrate high-level semantic abstractions with low-level geometric detail. Parallel structures, such as spatial pyramid pooling (SPP) \cite{spp}, atrous spatial pyramid pooling (ASPP) \cite{chen2018encoder}, and pyramid pooling modules (PPM) \cite{zhao2017pyramid}, capture contextual cues across different receptive fields. Sequential strategies like feature pyramid networks (FPN) \cite{lin2017feature} propagate hierarchical features in a top-down fashion, strengthening object representation at multiple scales.

Attention mechanisms further refine aggregated features. PraNet \cite{fan2020pranet} leverages reverse attention to progressively highlight boundary regions, while PolypPVT \cite{dong2021polyp} fuses CBAM \cite{woo2018cbam} with a PVTv2 backbone \cite{wang2022pvt} for fine-grained localization. CASCADE \cite{rahman2023medical} stacks channel and spatial attention, whereas EMCAD \cite{rahman2024emcad} utilizes large-kernel grouped attention gates to suppress irrelevant noise. More recently, $\text{MSA}^{2}\text{Net}$ \cite{kolahi2024msa} introduces multi-scale adaptive spatial attention to enhance spatial interaction and feature modulation. Yu et al. \cite{yu2026mmu} proposed  a novel skip connection module with DenseConvMixer, aligning and fusing multi-scale features to enhance global context, detail preservation, and segmentation accuracy. Shangguan et al. \cite{shangguan2026multi} proposed multi-stage fusion network MPFF-Net for stroke lesion detection. Collectively, these architectural and feature-level innovations including transformers, skip connection variants, and attention-driven enhancements have substantially advanced the accuracy, efficiency, and robustness of medical image segmentation systems.

\section{Proposed Method}

\begin{figure*}[t]
    \centering
    \includegraphics[ width=1.0\linewidth]{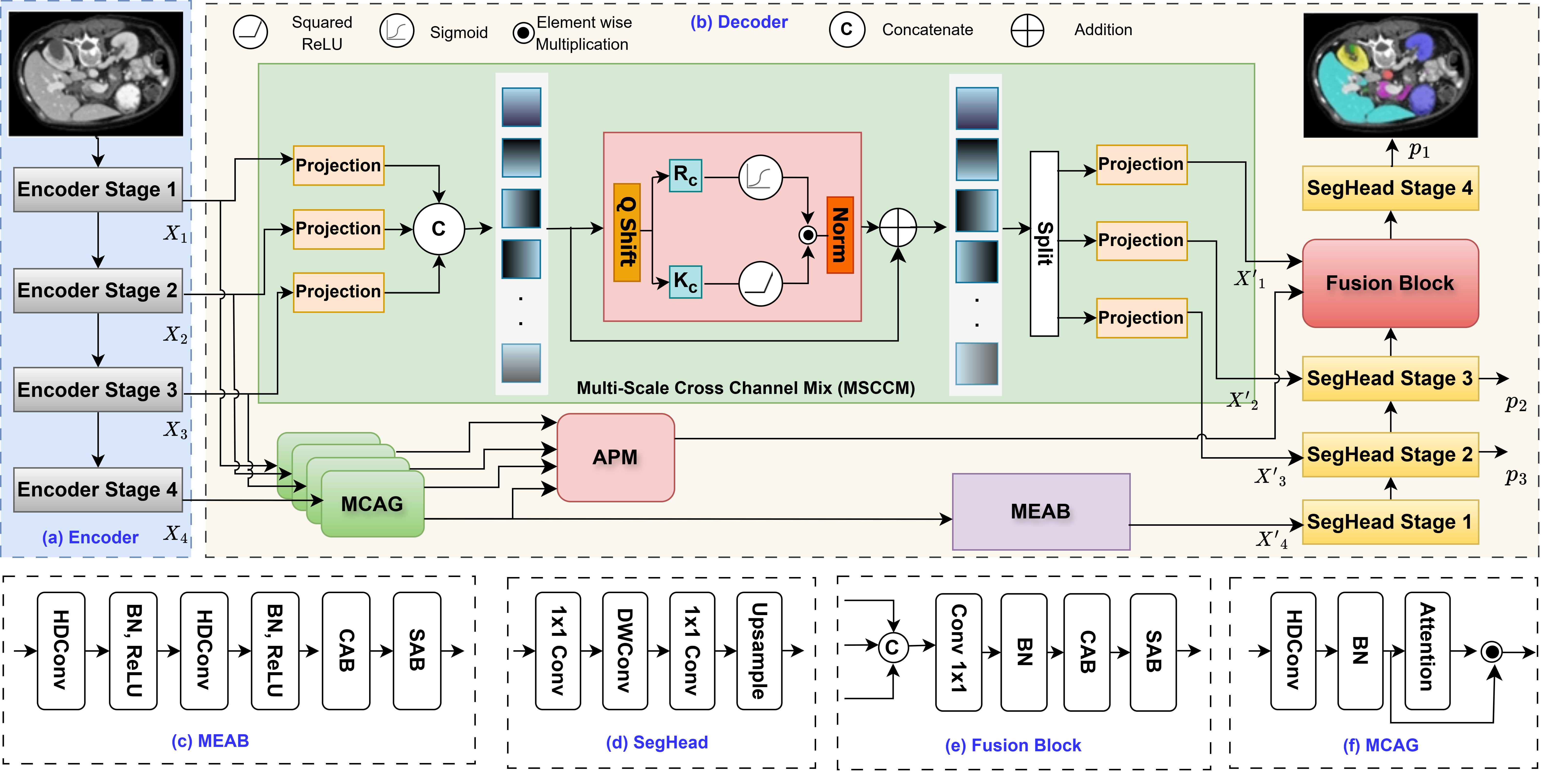}
    \caption{Overall architecture including hierarchical encoder with our proposed MAMCD decoder. (a) CNN or Transformer encoder with four stages (b) Overall MAMCD decoder (c) Multi-Dilated Enhanced Attention Block (MEAB) (d) Segmentation head at 4 stages (e) Fusion block including Channel attention block (CAB) and Spatial attention block (SAB) (f) Multi-dilated Contextual attention Gate (MCAG)}
    \label{fig1}
   \vspace{-0.3cm}
\end{figure*}

In this section, we introduce the overall design of the proposed \textbf{M}ulti-dilated contextual \textbf{A}ttention and \textbf{C}hannel \textbf{M}ixer \textbf{D}ecoding (MACMD) framework, as depicted in Fig. \ref{fig1}. Medical image segmentation requires both an encoder and a decoder, and we used two transformer-based architectures alongside our decoder in this study. The MACMD consists of four core modules designed to enhance receptive fields and contextual understanding, facilitating the accurate capture of anatomical structures. These modules function as advanced skip connections, merging the encoder’s high spatial resolution with the decoder’s semantically rich features for efficient multi-scale feature fusion across the encoding and decoding stages. As shown in Fig. \ref{fig1} (b), the MACMD comprises the Multi-dilated Contextual Attention Gate (MCAG), which highlights relevant regions within multi-scale features, and the Attention Pooling Modulation (APM), which compresses and reweights features into attention-enhanced pooled maps.  Both act as learnable attention-based modulators that increase the effective receptive field and selectively emphasize informative regions. The Multi Scale Cross-Channel Mixer (MSCCM) dynamically mixes features across channels through a recurrent mixing mechanism, while the Multi-dilated Enhanced Attention Block (MEAB) extracts high-quality features through multi-dilated convolution combined with spatial and channel attention. Specifically, pyramid features ($X_1$, $X_2$, $X_3$, $X_4$ in Fig. \ref{fig1} (a)) extracted from the encoder pass through these modules. MCAG+APM processes these features and integrates them into SegHead Stage 3. MSCCM utilizes features $X_1$, $X_2$, and $X_3$, with the output concatenated at SegHead Stages 2, 3, and the Fusion block. The MEAB refines the deepest feature ($X_4$) before forwarding it to SegHead Stage 1, ensuring enhanced feature representation. Further implementation details of each block are discussed in the following subsections. 

\subsection{Multi-dilated Contextual Attention Gate (MCAG)}
We employ a gating-based attention mechanism where the input feature map $x \in \mathbb{R}^{C \times H \times W}$ is passed through a multi-dilated convolutional transformation \cite{hu2024hdconv}, followed by a spatial attention mask. The mathematical formulation is as follow;
\begin{equation}
    x_1 = \text{BN}\left( \text{HDConv}(x) \right)
\end{equation}
\begin{equation}
    \alpha = \sigma \left( \text{BN}\left( \text{Conv}_{1 \times 1} \left( \text{ReLU}(x_1) \right) \right) \right)
\end{equation}
\begin{equation}
\hat{x} = x_1 \odot \alpha  
\end{equation}

\noindent Here, \( \odot\) represents element-wise multiplication and \(\alpha \in \mathbb{R}^{C \times H \times W}\) is applied to each channel of $x_i$. The HDConv layer applies four parallel dilated convolutions with different dilation rates (1, 2, 3, 5), splits their outputs into four channel groups, and combines them in a cyclic pattern to produce the final output. The above formulation allows the network to adaptively emphasize or suppress spatial regions of the feature map based on their learned importance, effectively refining the features using spatial context and attention.

\subsection{Attention Pooling Modulation (APM)}

\begin{figure}
    \centering
    \includegraphics[width=0.9\linewidth]{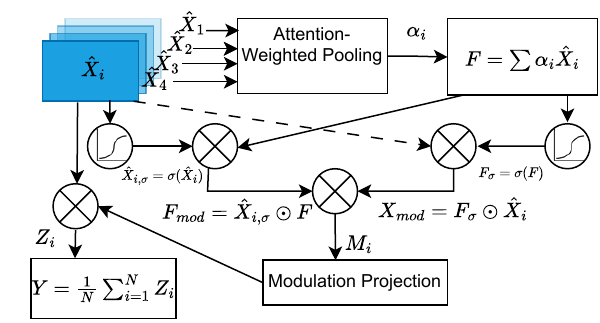}
    \caption{Our proposed APM module}
    \label{fig2}
\end{figure}

 Fig. \ref{fig2} presents the functional flow diagram of APM. The four feature maps $X_i \in \mathbb{R}^{C_i \times H_i \times W_i}$ obtained from MCAG are initially projected to a lower channel dimension (denoted as $C$) via $1\times1 $ convolution, followed by batch normalization (BN) and ReLU activation. This projection aligns features from different scales to a unified channel dimension across different scales, enabling consistent multi-scale fusion. Next, efficient upsampling is performed to match the spatial dimensions of the largest feature map $X_1$ ($H \times W$). Once all feature maps $\hat{X}_i \in \mathbb{R}^{C \times H \times W}$ are aligned, an attention score is computed, and the fused feature maps are obtained by applying the learned attention weights (Fig. \ref{fig2}), as expressed mathematically:

\begin{equation}
A_i = \text{Conv}_{1\times1}^{\text{attn2}}(\text{ReLU}(\text{Conv}_{1\times1}^{\text{attn1}}( \hat{X}_i))) \in \mathbb{R}^{1 \times H \times W}
\end{equation}
\begin{equation}
    \alpha_i = Softmax(A_i, dim=1)
\end{equation}
\begin{equation}
F = \sum_{i=1}^N \alpha_i \odot \hat{X}_i \in \mathbb{R}^{C \times H \times W}
\end{equation}

\noindent where, \( \text{Conv}_{1\times1}^{\text{attn1}}: \mathbb{R}^{ C_{\text{out}} \times H_{\text{target}} \times W_{\text{target}}} \to \mathbb{R}^{B \times \frac{C_{\text{out}}}{8} \times H_{\text{target}} \times W_{\text{target}}} \) and \\ 
\( \text{Conv}_{1\times1}^{\text{attn2}}: \mathbb{R}^{B \times \frac{C_{\text{out}}}{8} \times H_{\text{target}} \times W_{\text{target}}} \to \mathbb{R}^{B \times 1 \times H_{\text{target}} \times W_{\text{target}}} \)

\noindent To enhance the fusion, a bidirectional modulation is applied between local features $\hat{X}_i$ and the aggregated context $F$. The modulation weight $\gamma_i$ is computed by enriching each $\hat{X}_i$ with contextual information from $F$, and vice versa. This involves, (1) enhancing $\hat{X}_i$ with global context via a sigmoid-gated multiplication with $F$, forming $F_{mod}$, (2) Simultaneously, enriching $F$ with local detail from $\hat{X}_i$, forming $X_{mod}$. These modulated global and local features are fused via element-wise multiplication, followed by a $1\times1$ convolution, batch normalization, and ReLU, resulting in modulation weights $\gamma_i$ which merges fine-grained precision with contextual insights. \par  

\noindent The original projected feature is then modulated, and the final APM output $Y$ is obtained by averaging the modulated features as follow:
\begin{equation}
    Z_i = \gamma_i \odot \hat{X}_i, \quad Y = \frac{1}{N} \sum_{i=1}^N Z_i
\end{equation}
Through attention and modulation mechanisms in APM, the model adaptively focuses on relevant features, enhancing robustness and segmentation accuracy.

\subsection{Multi Scale Cross Channel Mix (MSCCM)}

Inspired by the Channel Mix module in VRWKV \cite{duan2024vision}, we introduced a MSCCM module designed to effectively capture and mix channel information from different stage of encoder. MSCCM enhances the representation of long-range contextual dependencies by leveraging global channel-wise interactions \cite{jiang2025rwkv}. Let the output feature maps from encoder Stage 1-3 be denoted as 
\(X_1 \in \mathbb{R}^{C_1 \times H_1 \times W_1}\), 
\(X_2 \in \mathbb{R}^{C_2 \times H_2 \times W_2} \), 
and \(X_3 \in \mathbb{R}^{C_3 \times H_3 \times W_3}\), where the spatial resolution decreases and the number of channels increases progressively across stages. To align the features for fusion, we upsample and project them to match the spatial size and channel dimension of \( X_1 \) using convolution:
\begin{equation}
    \tilde{F}_1 = \text{Conv}_{C_1}(F_1) 
\end{equation}
\begin{equation}
    \tilde{F}_i = \text{Conv}_{C_1}(\text{Upsample}(F_i)), \quad \text{where} \ i = 2, 3
\end{equation}
Thus, \( \tilde{X}_1, \tilde{X}_2, \tilde{X}_3 \in \mathbb{R}^{C_1 \times H_1 \times W_1} \). The features are then concatenated along the channel dimension. The concatenated feature map 
\(X_{\text{cat}} \in \mathbb{R}^{3C_1 \times H_1 \times W_1}\) is unfolded into a sequence of flattened patches \(X_{\text{unf}}\) and then we apply ChannelMix from VRWKV with residual connection to perform global feature fusion across the channel dimension:
\begin{equation}
X_{\text{mix}} = \text{ChannelMix}(\text{LayerNorm}(X_{\text{unf}})) + X_{\text{unf}} \in \mathbb{R}^{3C_1 \times N}
\end{equation}
\noindent where, N = $H_1 \times W_1$. The Channel Mix enhances feature fusion along channels using linear projections and gating. To capture 2D relationships, it integrates Q-Shift, interpolating tokens with neighbors in four directions. This minimally increases cost while expanding receptive fields, improving spatial coverage and efficiency in later layers for high-resolution image processing. This is followed by folding the features back to a 2D spatial format. We then split into three equal channel-wise parts. Each split is reshaped and passed through a convolution layer to restore its original spatial resolution and channel count. Finally, the refined features \( X'_1 \), \( X'_2 \), and \( X'_3 \) are forwarded to the fusion block, SegHead 3, and SegHead 2, respectively for further processing. 

\subsection{Multi-dilated Enhanced Attention Block (MEAB)}

This process is applied to the feature map of encoder Stage 4 after passing through the (MCAG). It begins with two sequential dilated convolutions (HDConv) utilizing multiple dilation rates (1, 2, 3, 5)  with a unique interleaved feature mixing strategy, which expand the receptive field without increasing the parameter count, thereby improving contextual understanding. These convolutions are followed by Batch Normalization (BN) and ReLU activation to stabilize training and enhance non-linearity. Subsequently, channel-wise attention is computed through average pooling and a two-layer MLP with ReLU and Sigmoid activations. Finally, both channel and spatial attention mechanisms refine feature selection by emphasizing relevant information and suppressing noise, leading to improved model efficiency and performance.
\subsection{Segmentation Head (SegHead)}
Our CNN-based segmentation heads (SegHead) block is constructed using a pointwise convolution layer followed by a $9 \times 9$ depthwise convolution (DW-Conv) layer. This is preceded by a point operation layer and an upsampling operation. 
From the SegHead 2, 3, and 4 we predict three output $p_1$, $p_2$ and $p_3$ respectively to calculate the loss. 
\subsection{Loss and outputs aggregation}
\textbf{Loss Function:} Following the approaches in \cite{milletari2016v} we utilized the weighted sum of cross entropy ($CE$) loss and $Dice$ loss between actual and prediction mask image formulated as follow; 
\begin{equation}
    \mathcal{L} = \alpha \, CE(p, y) + \beta \, Dice(p, y),
\end{equation}
where $\alpha$ and $\beta$ are the weight coefficient, $p$ and $y$ are prediction and actual mask. Additionally, the total loss is computed as the sum of the individual losses from the three predictions.
\begin{equation}
    \mathcal{L_{\text{total}}} = \sum_{i=1}^3 \mathcal{L}(p_i, y) . 
\end{equation}
\noindent \textbf{Aggregating Output Segmentation Maps:} We designate the prediction map $p_1$, produced at the SegHead stage 4 of our decoder, as the definitive segmentation map. To generate the final segmentation output, we apply a Sigmoid activation for binary segmentation tasks or a Softmax activation for multi-class segmentation tasks.

\begin{table}[!h]
\caption{Results of binary medical image segmentation for breast ultrasound (BUSI) and skin lesion segmentation (ISIC 2017).  The values are averaged over five runs. The best-performing result is highlighted in blue, while the second-best is marked in red.}
\label{table:binary}
\centering
\scriptsize
\resizebox{0.6\linewidth}{!}{%
\begin{tabular}{@{}lllll@{}}
\toprule
\multirow{2}{*}{Methods} & \multicolumn{2}{c}{BUSI} & \multicolumn{2}{c}{ISIC} \\ \cmidrule(l){2-5} 
                         & DSC         & ACC        & DSC         & ACC        \\ \cmidrule(r){1-1}
UNet \cite{ronneberger2015u}                     & 76.12       & 95.39      & 84.12       & 93.29      \\
AttnUnet \cite{oktay2018attention}                 & 77.64       & 95.94      & 83.88       & 95.38      \\
TransUnet \cite{chen2021transunet}                & 77.85       & 95.92      & 86.67       & 96.08      \\
Swin-Unet \cite{cao2022swin}               & 78.73       & 96.13      & 86.37       & 95.83      \\
UNeXt \cite{valanarasu2022unext}               & 74.37       & 95.85      & 85.38       & 93.44      \\
UCTransNet \cite{wang2022uctransnet}               & 79.04       & 96.39      & 86.89       & 94.96      \\
PVT-CASCADE \cite{rahman2023medical}         & 80.01       & 96.30      & 87.61       & 96.23      \\
PVT-EMCAD-B2 \cite{rahman2024emcad}        & 80.65       & 96.46      & 87.16       & 96.13      \\
\midrule
MaxViT-T-MACMD (\textbf{Ours})           & \textcolor{purple}{80.77}       & \textcolor{purple}{96.51}      & \textcolor{purple}{88.74}       & \textcolor{purple}{96.64}      \\
PVT-B2-MACMD (\textbf{Ours})             & \textcolor{blue}{81.67}       & \textcolor{blue}{96.63}      & \textcolor{blue}{89.82}       & \textcolor{blue}{96.99}      \\ \bottomrule
\end{tabular}%
}

\end{table}
\begin{table*}[!h]
\centering
\caption{Quantitative results for Synapse multi-organ segmentation. The top-performing result is marked in blue, while the second-best is shown in red. The reported metrics include parameters (in millions (M)), and FLOPs (in billions (G)). The DSC of individual organ is reported. Note GB: Gallbladder, KL and KR: Left and Right Kidney, PC: Pancreas, SP: Spleen, SM: Stomach }

\label{table:synapse}
\resizebox{1\textwidth}{!}{%
\begin{tabular}{@{}|l|ll|ll|llllllll|@{}}
\toprule
\multirow{2}{*}{Methods} &
  \multirow{2}{*}{Params} &
  \multirow{2}{*}{FLOPs} &
  \multicolumn{2}{c|}{Average} &
  \multirow{2}{*}{Aorta} &
  \multirow{2}{*}{GB} &
  \multirow{2}{*}{KL} &
  \multirow{2}{*}{KR} &
  \multirow{2}{*}{Liver} &
  \multirow{2}{*}{PC} &
  \multirow{2}{*}{SP} &
  \multirow{2}{*}{SM} \\\cmidrule(l){4-5}
                   &        &        & DSC $\uparrow$   & HD95 $\uparrow$     &       &       &       &       &       &       &       &       \\\cmidrule(l){1-13}
UNet \cite{ronneberger2015u}               & 34.53M & 65.53G & 70.11 & 44.69 & 84.00 & 56.70 & 72.41 & 62.64 & 86.98 & 48.73 & 81.48 & 67.96 \\
R50+AttnUNet   \cite{oktay2018attention} & 77.40M  & 30.70G  & 75.57 & 36.97 & 55.92 & 63.91 & 79.20 & 72.71 & 93.56 & 49.37 & 87.19 & 74.95 \\
TransUNet \cite{chen2021transunet}         & 105.28M & 24.73G  & 77.61 & 26.90 & 86.56 & 60.43 & 80.54 & 78.53 & 94.33 & 58.47 & 87.06 & 75.00 \\
SwinUNet \cite{cao2022swin}           & 41.38M  & 8.66G   & 77.58 & 27.32 & 81.76 & 65.95 & 82.32 & 79.22 & 93.73 & 53.81 & 88.04 & 75.79 \\
MT-UNet \cite{wang2022mixed}           & 79.08M  & 44.88G  & 78.59 & 26.59 & \textcolor{blue}{87.92} & 64.99 & 81.47 & 77.29 & 93.06 & 59.46 & 87.75 & 76.81 \\
HiFormer \cite{heidari2023hiformer}          & 29.52M  & 13.46G  & 80.69 & 19.14 & 87.03 & 68.61 & 84.23 & 78.37 & 94.07 & 60.77 & 90.44 & 82.03 \\
PVT-CASCADE \cite{rahman2023medical}       & 35.27M  & 6.28G   & 81.06 & 20.23 & 83.01 & 70.59 & 82.23 & 80.37 & 94.08 & 64.43 & 90.10 & \textcolor{blue}{83.69} \\
TransCASCADE \cite{rahman2023medical}      & 123.49M & -      & \textcolor{purple}{82.68} & \textcolor{purple}{17.34} & 86.63 & 68.48 & \textcolor{blue}{87.66} & \textcolor{blue}{84.56} & 94.43 & \textcolor{purple}{65.33} & 90.79 & \textcolor{purple}{83.52} \\
\midrule
MaxViT-T-MACMD (\textbf{Ours})     & 39.34M  & 7.34G   & 81.88 & 19.53 & 87.45 & \textcolor{blue}{73.94} & 82.54 & 80.06 & \textcolor{purple}{95.07} & 62.69 & \textcolor{purple}{91.08} & 82.17 \\
PVT-B2-MACMD (\textbf{Ours})      & 34.60M  & 6.02G   & \textcolor{blue}{83.27} & \textcolor{blue}{14.92} & \textcolor{purple}{87.85} & \textcolor{purple}{73.31} & \textcolor{purple}{86.28} & \textcolor{purple}{84.13} & \textcolor{blue}{95.34} & \textcolor{blue}{65.80} & \textcolor{blue}{92.78} & 80.68\\ \bottomrule
\end{tabular}%
}%
\end{table*}

\section{Experiment}
We evaluated the performance of two transformer-based architectures (MaxViT-T-MACMD and PVT-B2-MACMD) incorporating our proposed decoder. The performance has been compared against SOTA methods for both binary segmentation and multi-organ segmentation.\par  
\subsection{Datasets and Metrics} To evaluate our method, we conducted breast cancer segmentation using the BUSI dataset \cite{al2020dataset}, which consists of 780 breast ultrasound images (average resolution: $500\times500$ pixels), covering normal, benign, and malignant cases with corresponding segmentation masks. Our focus was on benign and malignant cases, totalling 647 images (437 benign, 210 malignant). Additionally, we addressed skin lesion segmentation using the ISIC 2017 \cite{codella2018skin} dataset, which includes 2000 pairs of dermoscopy images and their corresponding masks. For multi-organ segmentation, we employed the Synapse multi-organ dataset \cite{synapse2015ct}, comprising 30 abdominal CT scans with 3,779 axial contrast-enhanced slices. For multi-organ segmentation, Dice Similarity Coefficient (DSC) and Hausdorff Distance (HD95) are used as standard metrics, while accuracy (ACC) and DSC are utilized for binary classification compare with the state-of-the-art (SOTA) methods \cite{tragakis2023fully}.


\subsection{Implementation Details}
All experiments were performed on an NVIDIA A100 GPU with 40GB of memory.
Input resolutions were adjusted to $224\times224$ for Synapse and $256\times256$ for the BUSI and ISIC datasets. ImageNet-pretrained MaxViT-Tiny \cite{tu2022maxvit} and PVT-V2-B2 \cite{wang2022pvt} were utilized as encoders.
Model training was conducted using the AdamW optimizer with a learning rate of $1\times10^{-3}$, weight decay of $1\times10^{-4}$, and a CosineAnnealingLR scheduler. Batch sizes were set to 16 for Synapse and 24 for binary segmentation tasks. Models were trained for 150 epochs over five runs, except for Synapse, which underwent 300 epochs and was saved based on the best Dice score.
For the loss function, $\alpha = 0.4$ and $\beta = 0.6$ were used for Synapse, while for binary segmentation, $\alpha = \beta = 1$.
Dataset splits followed an 80:10:10 train-validation-test ratio for the BUSI dataset. In ISIC 2017, 1250 images were used for training, 150 for validation, and 600 for testing. Following TransUNet \cite{chen2021transunet}, we used the same 18 scans (2,212 axial slices) for training and 12 scans for validation in the Synapse dataset. All images underwent augmentation during training, as outlined in \cite{chen2021transunet}.

\section{Results and Discussion}
\begin{figure*}[h]
    \centering
    \includegraphics[width=1\linewidth]{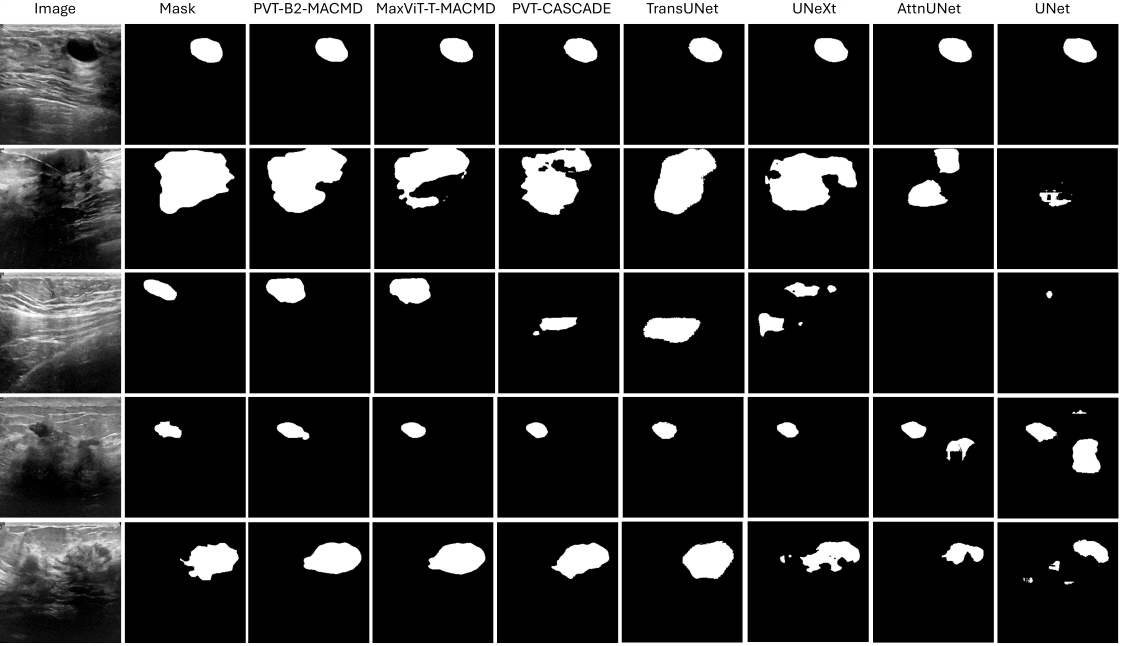}
    \caption{Visualization comparison between proposed decoder architectures with several methods on BUSI dataset}
    \label{fig_busi}
\end{figure*}

\subsection{Results of Binary Medical Image Segmentation}
Table \ref{table:binary} shows the binary segmentation results of the BUSI and ISIC datasets highlights the superior generalization and segmentation capability of our proposed methods MaxViT-T-MACMD and PVT-B2-MACMD. On the BUSI dataset, PVT-B2-MACMD achieves the highest Dice score of 81.67\% and accuracy of 96.63\%, surpassing prior best performers such as PVT-EMCAD-B2 (80.65 \% DSC) and UCTransNet (79.04\% DSC). Similarly, MaxViT-T-MACMD achieves a strong 80.77\% DSC. On the ISIC dataset, both models again outperform existing methods, with PVT-B2-MACMD achieving the best performance: 89.82\% DSC and 96.99\% accuracy, marking significant improvements over TransUNet (86.67\% DSC) and PVT-CASCADE (87.61\% DSC). Statistical analysis (p < 0.01) confirmed that our method significantly outperforms the implemented state-of-the-art approaches.The enhancements stem from the integration of attention-modulated multi-scale feature extraction and channel mixing. These results affirm the adaptability and precision of our architectures in handling diverse and challenging segmentation tasks across different medical imaging domains. 
Fig. \ref{fig_busi} shows the segmentation results of the BUSI dataset (a set of complex cases) across different methods. Our proposed models consistently align predicted boundaries with the ground truth, outperforming earlier approaches, particularly in complex scenarios. It achieves comprehensive nodule segmentation while minimizing disturbances. The model excels in diverse cases, including small, large, and challenging nodules resembling the background. It adeptly handles multiple nodules, eliminating irrelevant regions.\par
\begin{figure*}[h]
    \centering
  
    \includegraphics[width=1\linewidth]{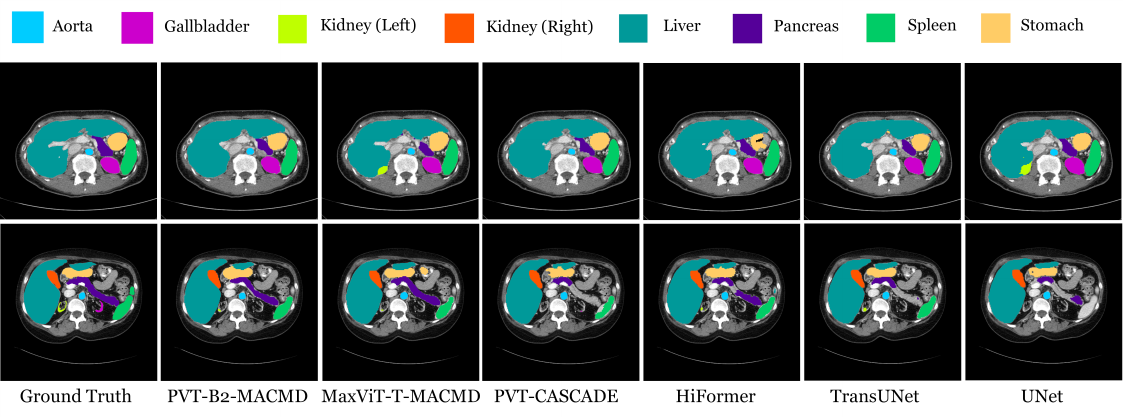}
      \caption{A Qualitative results and comparison with previous SOTA method on Synapse Multi-organ dataset. The visual results demonstrate that our method delivers more precise segmentation, particularly in challenging cases such as pancreas segmentation}
    \label{fig_synapse}
\end{figure*}

\subsection{Results of Multi-organ Segmentation} 
The proposed MaxViT-T-MACMD and PVT-B2-MACMD models demonstrate substantial improvements in DSC compared to existing segmentation methods, as shown in Table \ref{table:synapse}. Among them, PVT-B2-MACMD achieves the highest overall DSC of 83.27\%, representing an absolute improvement of 5.66\% over TransUNet (77.61\%) and 2.58\% over the previous best-performing method, TransCASCADE (82.68\%). Similarly, MaxViT-T-MACMD attains a DSC of 81.88\%, outperforming HiFormer (80.69\%) by 1.19\% and SwinUNet (77.58\%) by 4.3\%. On organ-wise analysis, PVT-B2-MACMD shows notable gains, especially in anatomically complex or highly variable organs. For instance, it achieves 65.80\% in pancreas segmentation, improving by 7.33\% over SwinUNet and 6.34\% over TransUNet. For spleen segmentation, it reaches 92.78\%, which is 6.80\% higher than UNet and 2.34\% higher than TransCASCADE. Across most organs, both proposed models consistently outperform baseline methods, validating the effectiveness of the attention-convolution hybrid architecture and the cross-scale feature fusion strategy for accurate and robust multi-organ segmentation. In addition to accuracy improvements, the proposed models also achieve significant efficiency gains as shown in Fig. \ref{fig_effi} . PVT-B2-MACMD reduces the number of parameters by approximately 67–72\% and the number of FLOPs by 75–91\% compared to transformer-based models such as TransUNet and TransCASCADE, while still delivering superior segmentation performance. MaxViT-T-MACMD also shows a marked reduction in computational complexity, with over 62\% fewer parameters and nearly 89\% fewer FLOPs than UNet, demonstrating excellent model efficiency. Fig. \ref{fig_synapse} presents qualitative segmentation results on the Synapse dataset, highlighting the effectiveness of our approach. The predictions closely follow the ground truth, preserving fine edge details and accurately delineating organ boundaries, even in challenging cases such as the pancreas.

\begin{table}

\centering
\caption{Ablation study on synapse dataset: effect of different modules of MACMD’s segmentation performance}
\label{table:ablataion}
\small
\resizebox{0.6\linewidth}{!}{%
\begin{tabular}{@{}cccll@{}}
\toprule
MCAG +APM & MSCCM & MEAB & Dice $\uparrow$  & HD95 $\downarrow$  \\ \midrule
\checkmark         &       &      & 80.59 & 16.33 \\
\checkmark         &       & \checkmark   & 81.14 & 18.14 \\
          & \checkmark     &      & 81.49 & 15.25 \\
\checkmark        & \checkmark     &      & 82.30 & 19.66 \\
          & \checkmark     & \checkmark    & 82.22 & 17.07 \\
\checkmark         & \checkmark     & \checkmark    & 83.27 & 14.92\\
\bottomrule
\end{tabular}%
}
\vspace{-0.3cm}
\end{table}
\subsection{Ablation Study}
We conducted an ablation study on the Synapse dataset to assess the impact of various module configurations on segmentation performance, as summarized in Table \ref{table:ablataion}. The study evaluates the individual and combined effects of MCAG + APM, MSCCM, and MEAB within our decoder design, using Dice score and HD95 as metrics. Incorporating MCAG + APM alone establishes a baseline Dice of 80.59\%. Adding MEAB slightly improves Dice to 81.14\%, though HD95 sees a minor increase. MSCCM alone  significantly enhances both metrics, achieving 81.49\% Dice and 15.25 HD95, underscoring its role in structural consistency. Combining MCAG + APM with MSCCM further boosts Dice to 82.30\%, while MSCCM with MEAB achieves 82.22\%, demonstrating complementary interactions. The complete integration of all three modules achieves the best performance, with 83.27\% Dice and 14.92 HD95, proving that each module contributes uniquely to feature fusion, boundary refinement, and contextual modeling. Additionally, retaining shallow skip connections is crucial, as shallow layers suffer greater information loss. The MSCCM module significantly improves segmentation, even when limited to one or two encoder inputs. Though it increases computational load, its global aggregation enhances accuracy. 
\subsection{Parameters Analysis}
\begin{figure}
    \centering
    \includegraphics[width=0.7\linewidth]{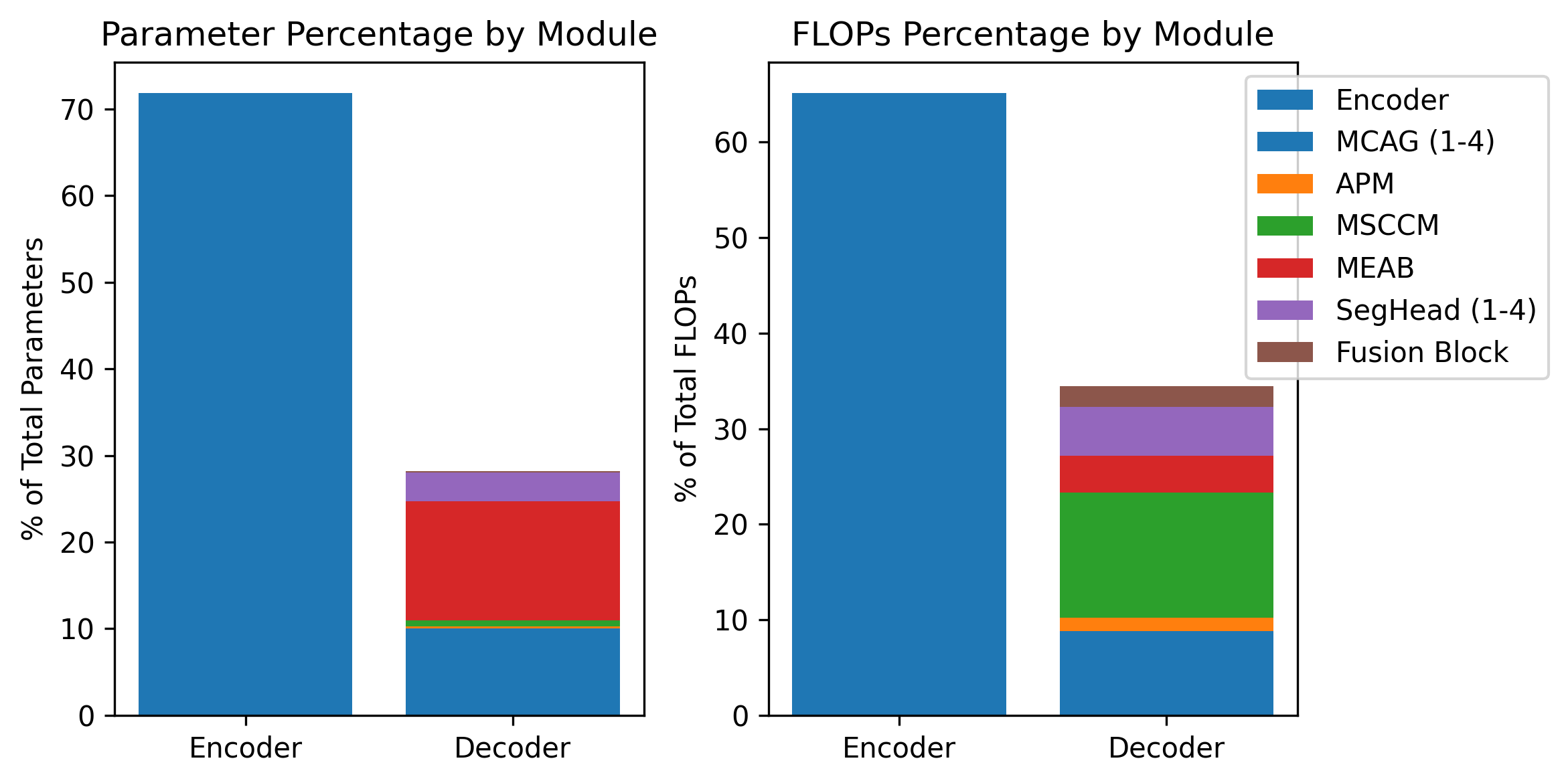}
    \caption{Parameter and FLOPs percentage of each module}
    \label{fig_params}
\end{figure}
The PVT-B2-MACMD model consists of 34.6 million parameters with a computational complexity of 6.02 GMac (Table \ref{tab_params} and Fig. \ref{fig_params}). It combines the PVT-V2-B2 encoder and the MACMD decoder, each contributing distinct computational loads. The encoder, forming the backbone of the model, accounts for 24.85 million parameters (71.82\%) and 3.92 GMac (65.10\%), serving as the primary feature extraction module. The decoder, composed of multiple sub-modules, has 9.75 million parameters (28.18\%) and 2.08 GMac (34.47\%). Among its components, the MCAG (1-4) contribute 3.47 million parameters (10.04\%) and 0.53 GMac (8.81\%), optimizing multi-scale feature interactions. MEAB is the most parameter-heavy in the decoder, with 4.76 million parameters (13.75\%), though its computational cost remains relatively low at 0.232 GMac (3.85\%). The MSCCM is the most computationally intensive, requiring 0.788 GMac (13.07\%), despite its lightweight parameter count of 0.252 million (0.73\%), highlighting dense feature modulation operations. The segmentation heads (SegHead 1-4) balance parameter efficiency (1.154 million parameters, 3.34\%) with computational demands (0.311 GMac, 5.17\%). The APM is highly efficient with only 0.071 million parameters (0.205\%) and 0.085 GMac (1.42\%), facilitating cross-level attention pooling. Lastly, the Fusion Block, the most compact component, integrates features with minimal overhead, contributing 0.043 million parameters (0.125\%) and 0.130 GMac (2.16\%). Overall, the model is optimized for balancing parameter efficiency and computational intensity, with MSCCM handling dense modulation, while MCAG and MEAB lead in parameter usage for enhanced attention-driven improvements.
\begin{figure}
    \centering
    \includegraphics[width=1\linewidth]{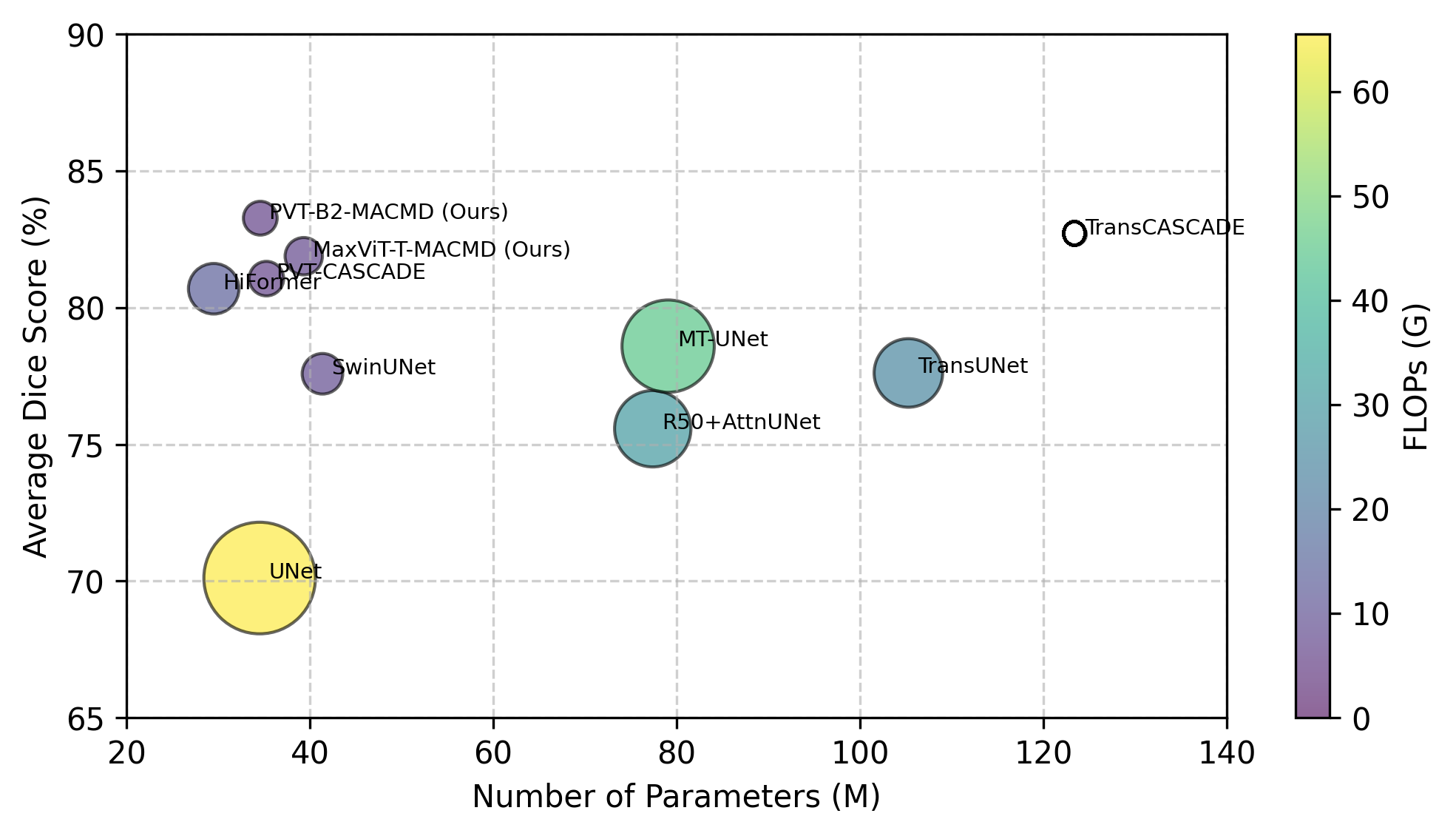}
    \caption{Comparison of methods in terms of average Dice score versus number of parameters. Circle size indicates computational cost measured in FLOPs}
    \label{fig_effi}
\end{figure}

\begin{table}[!h]
\centering
\label{tab_params}
\small
\caption{Details about the number of parameters in each module of MACMD decoder}
\resizebox{1\textwidth}{!}{%
\begin{tabular}{@{}lllll@{}}
\toprule
Module                     & Parameters (M) & \% of Total Params & FLOPs (GMac) & \% of Total FLOPs \\ \midrule
\textbf{Encoder (PVT-V2-B2)} & 24.85          & 71.815            & 3.92         & 65.104           \\
\textbf{Decoder (MACMD)}   & 9.75           & 28.18\%            & 2.08         & 34.47\%           \\ \midrule
MCAG   (1-4)               & 3.469          & 10.036\%           & 0.530        & 8.807\%           \\
APM                        & 0.071          & 0.205\%            & 0.085        & 1.418\%           \\
MSCCM                      & 0.252          & 0.727\%            & 0.788        & 13.073\%          \\
MEAB                       & 4.760          & 13.749\%           & 0.232        & 3.845\%           \\
SegHead   (1-4)            & 1.154          & 3.337\%            & 0.311        & 5.174\%           \\
Fusion   Block             & 0.043          & 0.125\%            & 0.130        & 2.157\%           \\ \bottomrule
\end{tabular}%
}

\end{table}

\subsection{GradCAM Visualization}
\begin{figure}[!h]
    \centering
    \includegraphics[width=1\linewidth]{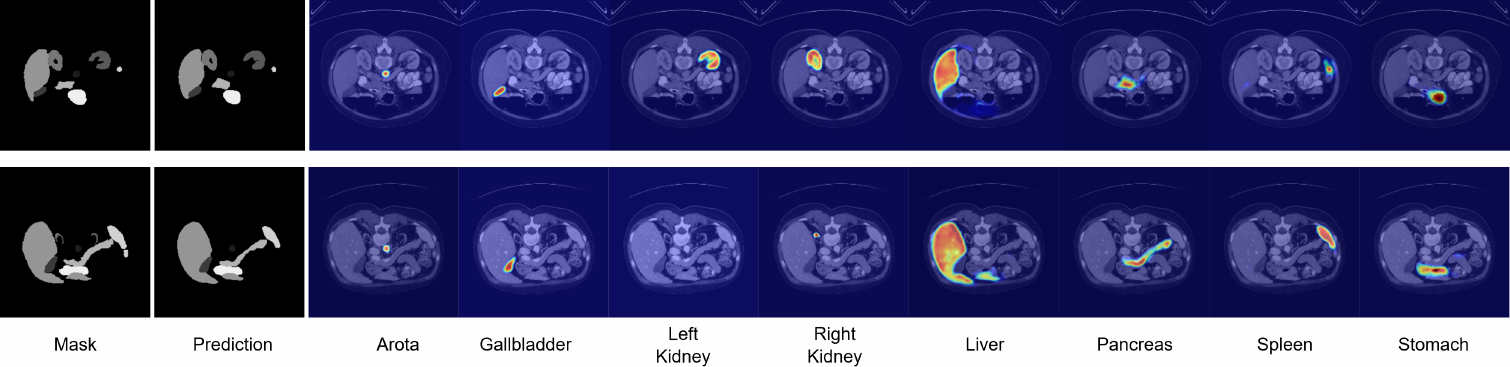}
    \caption{Feature visualization of two test samples from the Synapse dataset using our MACMD-based decoder}
    \label{fig:grad_cam}
\end{figure}

Figure \ref{fig:grad_cam} illustrates a Grad-CAM-based attention analysis of the proposed PVT-B2-MACMD model, particularly focusing on the MACMD decoder, applied to the Synapse multi-organ CT segmentation dataset. The image is structured into two main columns representing different test cases and ten rows that depict different organs and segmentation outputs. The top two rows show the ground truth segmentation masks and the model's predicted outputs. A close visual alignment between these two indicates the model’s strong segmentation performance across various organs. The subsequent rows display Grad-CAM heatmaps overlaid on CT slices for individual organs, including the aorta, gallbladder, kidneys, liver, pancreas, spleen, and stomach. These heatmaps represent the spatial attention captured by the MACMD decoder during prediction. High activation areas, shown in red or yellow, indicate regions the model found important for organ identification, while blue or darker regions reflect low attention. The Grad-CAM visualizations highlight the decoder’s ability to selectively attend to anatomically relevant regions, thereby enhancing interpretability. For smaller organs such as the aorta, gallbladder, pancreas, and stomach, the attention is highly localized, demonstrating the model’s capacity to capture fine-grained details and maintain high spatial resolution. In contrast, for larger organs like the liver, spleen, and kidneys, the model distributes attention more broadly, indicating its effectiveness in modeling global context and long-range dependencies. This dual capability is a direct result of the MACMD decoder’s hybrid design, which integrates multi-scale attention and spatially adaptive feature aggregation through hierarchical dilated convolutions, attention-driven modulation, and a cross channel-mixing module. Overall, the Grad-CAM maps confirm that the MACMD decoder balances local and global feature learning, allowing PVT-B2-MACMD to perform robust and reliable segmentation across organs of varying size and complexity. The model not only delivers high accuracy but also offers transparent decision-making, making it a promising architecture for clinical applications in medical image analysis.

Despite the effectiveness of the MACMD decoder in capturing both small and large organs, Grad-CAM visualizations reveal certain limitations. For instance, in the case of the stomach (as shown in Figure \ref{fig:grad_cam}), the attention map shows mild activation extending beyond the actual organ boundary, indicating the presence of diffuse or misplaced attention in surrounding regions. 

In the Synapse dataset, individual CT slices may occasionally lack corresponding label annotations or exhibit inaccurate segmentations. When employing attention-based models for 2D image segmentation, such inconsistencies can adversely impact segmentation quality, particularly in neighboring slices where inter-slice context is important. Furthermore, many existing segmentation architectures operate on 3D volumes in a slice-by-slice fashion, inherently ignoring spatial correlations across adjacent slices. This limitation reduces the model’s ability to leverage volumetric context, making it more prone to errors in anatomically ambiguous regions where organ boundaries are less distinct.

Notably, organs such as the pancreas and spleen, which often appear with low contrast and irregular shapes, demonstrate less focused and sometimes misaligned attention maps. These observations highlight the need for improved attention calibration, enhanced boundary refinement, and the integration of stronger spatial priors to boost segmentation accuracy in anatomically challenging regions.

\section{Conclusion}

This paper introduces MACMD, a novel approach that enhances skip connections for local-global feature interaction in encoder-decoder architectures. MACMD integrates attention enhancement at each encoder stage, pooling modulation for cross-scale integration in shallow layers, and channel mixing across different feature scales to refine local and global dependencies. Extensive experiments on binary and multi-organ segmentation tasks demonstrate its superior performance when integrating two encoders. PVT-B2-MACMD achieves the highest DSC (83.27\%) and the lowest HD95 (14.92), with lower GFLOPs, outperforming both CNN and transformer-based baselines. MaxViT-T-MACMD consistently surpasses CNN-based models while closely competing with top transformer architectures, maintaining efficiency. Our architectures strike an optimal balance between accuracy and computational efficiency, making them ideal for high-performance medical image segmentation. The MACMD decoder effectively addresses key limitations in current models, providing a practical and scalable solution for precise segmentation in clinical applications.


\end{document}